\documentclass[letterpaper, 10 pt, conference]{ieeeconf}  

\IEEEoverridecommandlockouts                              

\usepackage{graphicx}
\usepackage{graphics} 
\usepackage{amsmath} 
\usepackage{lipsum}  
\usepackage{algorithm}
\usepackage[noend]{algpseudocode}
\usepackage{bm}
\usepackage{breqn} 
\usepackage{mcite}
\usepackage{courier}
\usepackage{todonotes}
\usepackage{amsfonts}
\usepackage{subfigure}
\usepackage{nicefrac}
\usepackage[export]{adjustbox}
\usepackage{color,soul}
\usepackage{amssymb}

\makeatletter
\renewcommand*\env@matrix[1][\arraystretch]{%
  \edef\arraystretch{#1}%
  \hskip -\arraycolsep
  \let\@ifnextchar\new@ifnextchar
  \array{*\c@MaxMatrixCols c}}
\makeatother
\newcommand{\matr}[1]{\mathbf{#1}}

\algdef{SE}[DOWHILE]{Do}{doWhile}{\algorithmicdo}[1]{\algorithmicwhile\ #1}%

\DeclareMathOperator*{\argmin}{arg\,min}
\title{\LARGE \bf
Learning Robust Manipulation Skills with Guided Policy Search via Generative Motor Reflexes
}

\author{Philipp Ennen$^{1}$, Pia Bresenitz$^{1}$, Rene Vossen$^{1}$, Frank Hees$^{1}$%
\thanks{$^{1}$Philipp Ennen, Pia Bresenitz, Rene Vossen, Frank Hees are with the Cybernetics Lab IMA \& IfU at RWTH Aachen University, Germany.}
\thanks{\tt\small philipp.ennen@rwth-aachen.de}%
}

\begin{document}
\maketitle
\thispagestyle{empty}
\pagestyle{empty}

\begin{abstract}
Guided Policy Search enables robots to learn control policies for complex manipulation tasks efficiently. Therein, the control policies are represented as high-dimensional neural networks which derive robot actions based on states. 
However, due to the small number of real-world trajectory samples in Guided Policy Search, the resulting neural networks are only robust in the neighbourhood of the trajectory distribution explored by real-world interactions.
In this paper, we present a new policy representation called \textit{Generative Motor Reflexes}, which is able to generate robust actions over a broader state space compared to previous methods. In contrast to prior state-action policies, Generative Motor Reflexes map states to parameters for a state-dependent motor reflex, which is then used to derive actions.
Robustness is achieved by generating similar motor reflexes for many states. 
We evaluate the presented method in simulated and real-world manipulation tasks, including contact-rich peg-in-hole tasks. Using these evaluation tasks, we show that policies represented as Generative Motor Reflexes lead to robust manipulation skills also outside the explored trajectory distribution with less training needs compared to previous methods.
\end{abstract}

\section{Introduction}
\textit{Guided Policy Search} (GPS) is a model-based reinforcement learning approach that allows for learning manipulation skills in a continuous state and action space \cite{c2}. 
In reinforcement learning, manipulation skills are represented by policies which equal a transfer function to compute state-dependent actions.
Guided Policy Search is known as a sample-efficient approach which learns a neural network policy in two phases. First, a trajectory-centric model-based reinforcement learning is used to train local trajectories for different start and goal conditions. Then, these local trajectories serve as training data for supervised learning of a neural network policy. 

\begin{figure}
	\centering
	\includegraphics[width=0.41\textwidth, trim=0 2 0 2,clip]{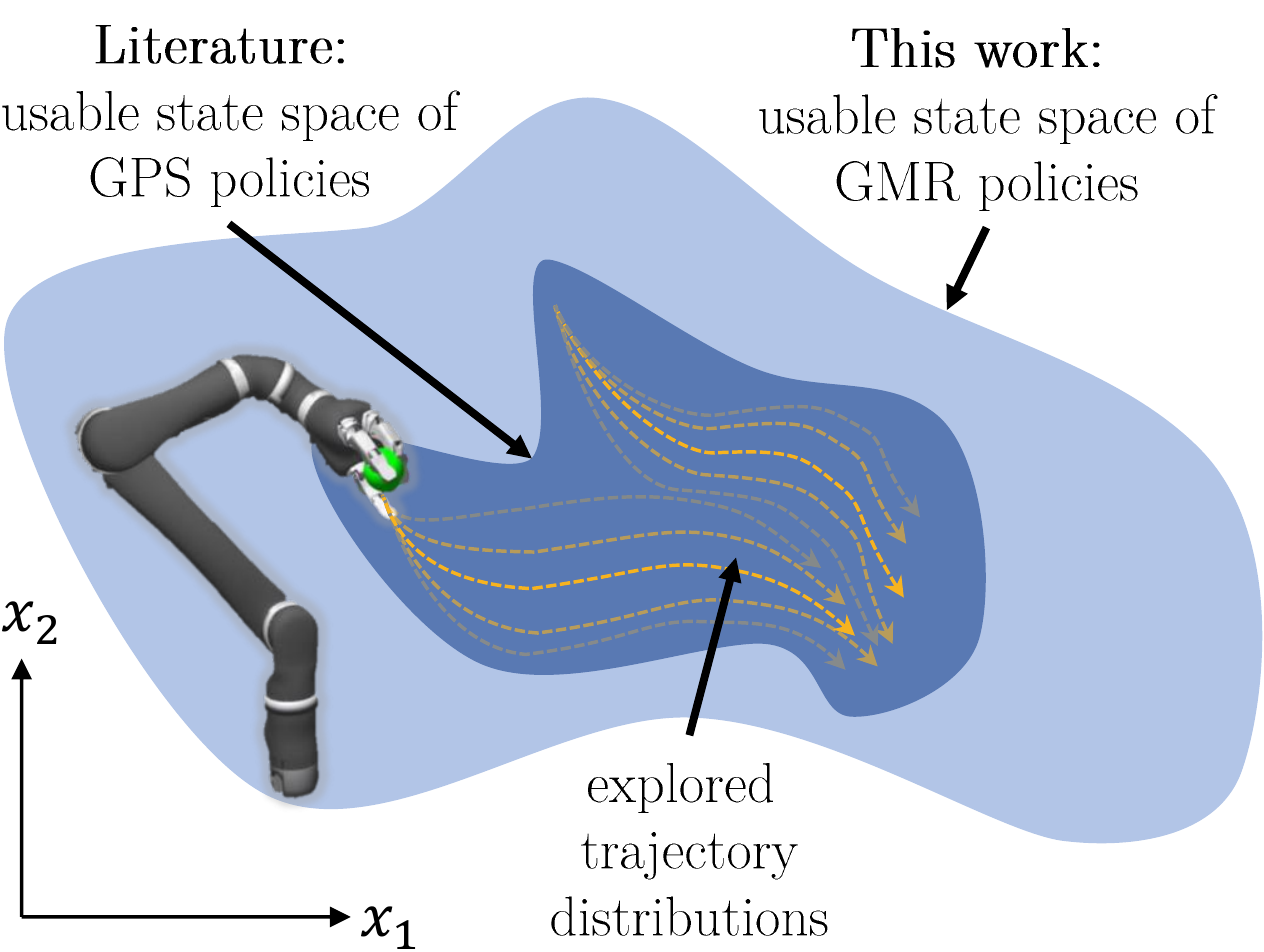}
	\caption{The aim of our work is to improve the robustness of high-dimensional neural network policies trained by Guided Policy Search (GPS) outside the explored state space (yellow trajectory distribution). Typically, policies learned with Guided Policy Search are only robust in the local neighbourhood of the explored state space (dark blue state space). We introduce Generative Motor Reflexes (GMR) that yields robustness in an increased state space (light blue state space).}
    \label{fig:idea}
\end{figure}

\noindent An essential requirement for many applications in robotics is a predictable and reliable behaviour of policies. For this reason, the policies need to be robust against unmodeled and unexpected disturbances in the state space. However, in GPS the resulting neural network policies are only robust in a small area around the explored state space since the training data for the neural network only contains a moderate amount of trajectories.
In case of unexpected state disturbances during test time, a policy easily produces unstable robot behaviour. For that reason, our work aims to reach robustness, even outside the distribution of the explored trajectories, cf. Figure \ref{fig:idea}.

\begin{figure*}[bt]
	\centering
	\includegraphics[width=0.85\textwidth, trim=0 0 0 9,clip]{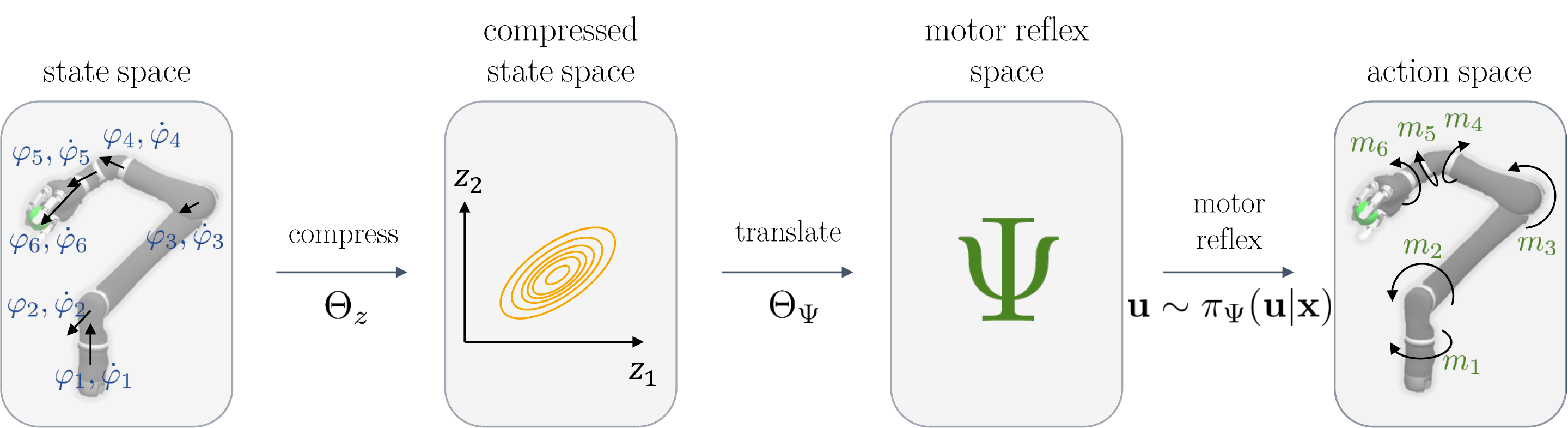}
    \caption{Overview of the proposed policy representation. From left to right: the state space describes the robot configuration which is in our work the six joint angles $\varphi_i$ and velocities $\dot{\varphi}_i$. By using a neural network with weights $\Theta_z$, the state space is compressed to a lower dimensional representation $\matr{z}$. Consequently, the compressed representation of many state is inside or very close to the trained distribution.
    Then, the compressed state is translated to motor reflex parameters $\Psi$ using a neural network with weights $\Theta_{\Psi}$. The motor reflex parameters form a motion pattern $\pi_{\Psi}(\matr{u}|\matr{x})$, which subsequently leads to actions $\matr{u}$ to be executed. The robot used in this paper is a simulated and real Kinova Jaco 2, which is controlled by six joint torques $m_i$. Supplementary video:  \texttt{https://philippente.github.io/pub/GMR.html}}
    	\label{fig:overview}
\end{figure*}

\noindent We introduce a novel and robust policy representation called \textit{Generative Motor Reflexes} (GMR), cf. Figure \ref{fig:overview}. This policy works as a two-step approach deriving an action. In the first step, a neural network predicts parameters of a motor reflex for a given state. In the second step, the parametrized motor reflex is used to derive stabilizing actions. A motor reflex is a simple one time step motion pattern containing a stabilizing state feedback that shares the same parameters for a region of input states. 

\noindent Generative Motor Reflexes increase the robustness if they fulfill two criteria: 
(i) the generated motor reflex parameters are drawn from the distribution used for exploring the state space and (ii) the generated motor reflex has stabilizing properties. Afterwards, during test time the stabilizing properties of a motor reflex are forcing the robot back to the explored state space. 
Compared to prior state-action policies, 
our approach results in a constrained parameter space of the motor reflex and as a consequence state disturbances are not yielding a self-resonating error. 
To summarize, our main contributions are:
\begin{enumerate}
 	\item A novel neural network policy representation called Generative Motor Reflex that uses motor reflexes as an intermediate step for deriving stabilizing actions and 
 	\item an adapted version of Guided Policy Search, which allows for training a Generative Motor Reflex policy. 
\end{enumerate}
We derive our approach by reviewing the related work in the field of reinforcement learning and in particular robust reinforcement learning. Then, we explain our Generative Motor Reflex policy in detail and present a training algorithm based on Guided Policy Search. Finally, we evaluate our approach on simulated and real-world experiments in the continuous control domain.

\section{Related work}
Reinforcement learning has been applied successfully in simulated and real-world robotic manipulation \cite{c7,c8,c9,c10,c11,c12}, locomotion \cite{c5,c6} and autonomous vehicles \cite{c13}. Many of the demonstrated scenarios used tailored policy representations or discretized action spaces. Due to the difficulties in training continuous high-dimensional policies efficiently, the parameter space often contains less than 100 parameters \cite{c1}. Although notable research has been conducted 
in learning methods for high-dimensional neural network policies, the methods only been developed 
to the point where 
they could be applied in the continuous control domain with moderate state and action spaces, which is typically around seven degrees-of-freedom \cite{c14}. Neural networks often lack robustness, so that today’s applications are limited, in particular in continuously controlled systems. Thus, recent research addresses robustness in two ways: robust policy representations and robust reinforcement learning, which are both reviewed in the following.  
\subsection{Robust Policy Representations}
One approach for robust policies is to use attractor systems as a policy representation. \textit{Dynamic Movement Primitives} (DMP) belong to this class, wherein robustness is obtained 
by a nonlinear, error-minimizing dynamic system \cite{c14.1}. 
Dynamic Movement Primitives are used to generate variations of an original movement, whereby the shape of the movement is represented by a mixture of Gaussian basis functions. Although DMPs have demonstrated to be successful in many applications, their generalization ability is limited. In recent work, DMPs have been extended to adapt to unknown sensor measurements via a feedback model learned from demonstration \cite{c14.1.2} \cite{c14.1.3}.

\noindent Aside from attractor systems, many different general-purpose policy representations have been developed. \textit{Probabilistic Movement Primitives} (ProMP), \textit{Gaussian Mixture Models} (GMM) and \textit{Hidden Markov Models} (HMM) are approaches that are modeling a trajectory distribution from stochastic movements \cite{c14.2, c14.2GMM, 14.2HMM}. In recent work, \textit{neural-network-based end-to-end sensorimotor policies} became very popular for their generalization abilities \cite{14.2ANN}. But these representations do not provide robustness by design. Typically, robustness of such general-purpose representations, i.e. ProMP, GMM, HMM and ANN, is only achieved, if the state disturbances are inside the trajectory distribution explored by the robot during training time. 
This motivated the development of several robust reinforcement learning methods. 
\subsection{Robust Reinforcement Learning}
In \textit{Robust Reinforcement Learning} (RRL), robustness is obtained 
by explicitly taking into account input disturbances and modeling 
errors \cite{c14.25}. 
In order to learn a robust policy, a control agent is trained in the presence of a disturbance agent, who tries to generate the worst possible disturbance. 
\textit{Robust Adversarial Reinforcement Learning} (RARL) uses an adversarial neural network for the disturbance agent \cite{c14.3}. 
The adversarial is trained jointly with the control agent and learns an optimal destabilization policy. 

\noindent A related approach is \textit{Training with Adversarial Attacks} (TAA) \cite{c14.4}. Instead of training an adversarial, knowledge about the Q-value is exploited to craft optimal adversarial attacks. The Q-value describes the expected cost-to-go of an agent given a state and an action. In TAA, adversarial attacks are seen as disturbances that result from the worst possible action, which corresponds to the smallest Q-value in a state. 

\noindent However, in RRL, robustness is reached by exploring those states, which could potentially lead to unstable behaviour. In contrast, GPS trains a policy by imitating multiple trajectory distributions learned by local, trajectory-centric reinforcement learning for different start and goal conditions \cite{c2}. As a result, neural network policies are only robust in the neighbourhood of the explored trajectory distributions. The extension of GPS for RRL would require a more widely exploration decreasing the sample-efficiency.

\noindent In the following we present Generative Motor Reflexes that are robust in a much broader state space compared to prior neural network policies without further exploration needs. 
\section{Generative Motor Reflex}
\subsection{Preliminaries in Guided Policy Search}
In Guided Policy Search, neural network policies $\pi(\matr{u}|\matr{x})$ distributed over actions $\matr{u}$ and conditioned on the state $\matr{x} \in \mathcal{X}$ are trained by a trajectory-centric model-based reinforcement learning method combined with an alternating optimization procedure, cf. Algorithm \ref{alg:mdgps_algorithm} (C- and S-step). First, the robot samples a mini-batch $\mathcal{T}=\{\tau_i\}$ of trajectories  $\tau_i =\{\matr{x}_0,\matr{u}_0,...,\matr{x}_T,\matr{u}_T\}_i$ for multiple start and goal conditions $i$ and fits these trajectories to a time-varying Gaussian dynamic model $p_i(\matr{x}_{t+1}|\matr{x}_t, \matr{u}_t)$. This Gaussian dynamic model is represented by $$p(\matr{x}_{t+1}|\matr{x}_t, \matr{u}_t) = \mathcal{N}(f_{\matr{xu}t}[\matr{x}_t; \matr{u}_t]^{\text{T}}+f_{\matr{c}t}, \matr{F}_{t})$$ 
with the fitted time-varying matrix $f_{\matr{xu}t}$, the vector $f_{\matr{c}t}$ and the covariance $\matr{F}_{t}$, where the subscripts denote differentiation with respect to the vector $[\matr{x}_t; \matr{u}_t]$.

\noindent Now, the goal of the C-step is to compute improved local policies $p_i(\matr{u}_t|\matr{x}_t)$ by minimizing a quadratic cost function $$J(\tau)=\sum_{t=0}^TE_{p_i(\tau)} [l(\matr{x}_t,\matr{u}_t)]$$ 
of the form $$l(\matr{x}_t, \matr{u}_t)=\frac{1}{2}[\matr{x}_t;\matr{u}_t]^Tl_{\matr{x}\matr{u},\matr{x}\matr{u}t}[\matr{x}_t;\matr{u}_t]+[\matr{x}_t;\matr{u}_t]^Tl_{\matr{x}\matr{u}t}+\text{const} 
$$
 while the change of the local policies is bounded by a trust-region constraint. The constraint used in GPS is the \textit{Kullback-Leibler} (KL) 
 divergence to the previous policy distribution $D_{\text{KL}}(p_{i}(\tau)||\pi(\tau))\leq\epsilon$, where the policy distribution is computed by $$p_i(\tau)=p_i(\matr{x}_0)\prod_{t=0}^{T-1}p_i(\matr{x}_{t+1}|\matr{x}_t, \matr{u}_t)p_i(\matr{u}_t|\matr{x}_t).$$ 

\begin{algorithm}[b]
\caption{Guided Policy Search}\label{alg:mdgps_algorithm}
\begin{algorithmic}[1]
\For{$n$-iteration $n=1$ to $N$}
\State Generate samples $\mathcal{T}=\{\tau_i\}$ by running $p_i$ or $\pi$
\State Fit Gaussian dynamics $p_i(\matr{x}_{t+1}|\matr{x}_t, \matr{u}_t)$
\State \textbf{C-step}: $p_{i} \leftarrow \argmin_{p_i} J(\tau)$ \Statex \hspace*{10mm}  s.t. $D_{\text{KL}}(p_{i}(\tau)||\pi(\tau))\leq\epsilon$
\State \textbf{S-step}: $\pi \leftarrow \argmin_{\bm{\Theta}} \sum_{i} D_{\text{KL}}(\pi(\tau)||p_{i}(\tau))$
\EndFor
\end{algorithmic}
\end{algorithm}
\noindent Afterwards, the weights of the neural network policy $\pi$ are trained to imitate the trajectory distribution $p_i (\tau)$ using supervised learning (S-step).\\
\subsection{Problem Formulation}
In Guided Policy Search the robot explores a state distribution $p(\mathcal{X}_{\text{train}})$, which is a subset of the state space $\mathcal{X}$ \cite{c2}. In this explored state distribution $p(\mathcal{X}_{\text{train}})$, policies trained by Guided Policy Search are typically robust against state noise. However, robustness outside of $p(\mathcal{X}_{\text{train}})$ is usually not obtained.
Therefore, we aim to learn a stochastic policy 
\begin{align*}
\matr{u} \sim \pi(\matr{u}|\matr{x} + \bm{\varepsilon})
\end{align*}
that is robust against state noise $\bm{\varepsilon}$ in a broader state space than $p(\mathcal{X}_{\text{train}})$. 

\begin{figure}[bt]
	\centering
		\includegraphics[width=0.44\textwidth, trim=0 35 0 10,clip]{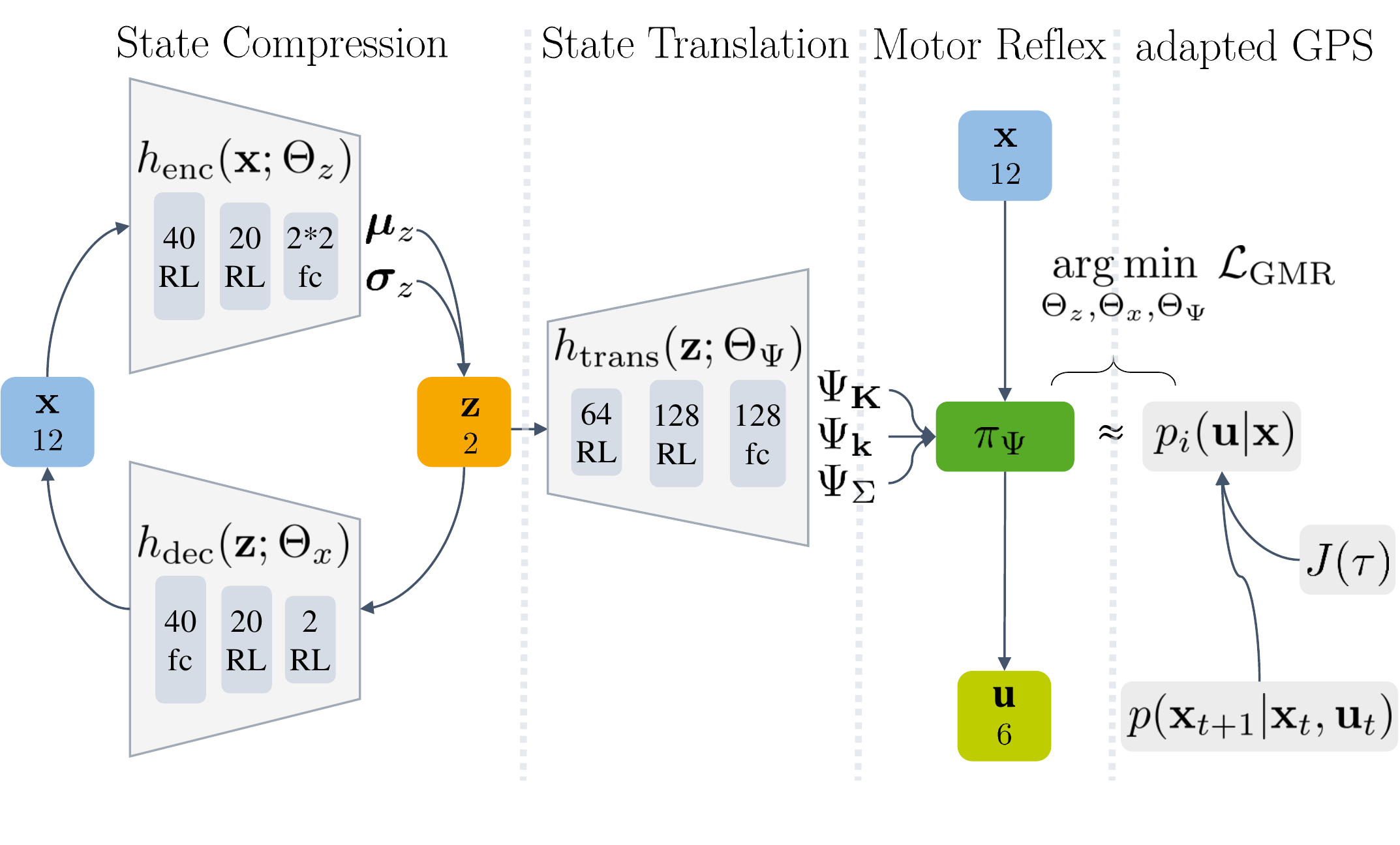}
	\caption{Generative Motor Reflex (GMR) policy and training: The state compression uses a variational autoencoder to encode the state $\matr{x}$ to a latent representation $\matr{z}$ using $h_{\text{enc}}$ and reconstruct it using $h_{\text{dec}}$. The translation model $h_{\text{trans}}$ transforms $\matr{z}$ to motor reflex parameters $\Psi_{\matr{K}}, \Psi_{\matr{k}}$ and $\Psi_{\matr{\Sigma}}$ which form a stochastic motor reflex $\matr{u} \sim \pi_{\Psi}(\matr{u}|\matr{x})$. The GMR weights $\bm{\Theta}_z, \bm{\Theta}_x$ and $\bm{\Theta}_{\Psi}$ are adapted to imitate the local policies $p_i(\matr{u}|\matr{x})$ by minimizing the loss term $\mathcal{L}_{\text{GMR}}$. For this, the local policies are derived by trajectory optimization using the dynamics $p(\matr{x}_{t+1}|\matr{x}_t, \matr{u}_t)$ and loss term $J(\tau)$. For the GMR model we set the hyperparameters to the stated number of neurons. Note: RL stands for leaky rectified linear units, fc for fully connected linear layers.}
    	\label{fig:gcm_model}
\end{figure}

\noindent For this purpose, the Generative Motor Reflex policy calculates the action via a compressed state space $\bm{z}$, generated motor reflex parameters $\bm{\Psi}$ and a state-dependent motor reflex $\pi_{\Psi}(\matr{u}|\matr{x})$ (cf. Figure \ref{fig:gcm_model}).
In this policy, the compressed state $\bm{z}$ is determined by a neural network using weights $\bm{\Theta}_z$ and the motor reflex parameters $\bm{\Psi}$ are determined by a neural network using weights $\bm{\Theta}_\Psi$. The motor reflex parameters $\bm{\Psi} = [\Psi_{\matr{K}}; \Psi_{\matr{k}};\Psi_{\matr{\Sigma}}]$ form a motor reflex $\pi_{\Psi}(\matr{u}|\matr{x})$, which is a stochastic motion pattern for one time step that computes actions using the framework of linear Gaussian controllers. 
\begin{align}
\pi_{\Psi}(\matr{u}|\matr{x}): \quad \matr{u} \sim \mathcal{N}(\Psi_{\matr{K}} \matr{x} + \Psi_{\matr{k}},\Psi_{\bm{\Sigma}})
\label{eq:lin_gaus_ctr}
\end{align}
A sequence of linear Gaussian controller of this form can be thought of as a trajectory together with a stabilizing feedback by $\matr{\Psi}_{\matr{K}}$. Current policy representations in GPS do not explicitly consider this feedback so far.


\noindent Then, during training time, the robot explores a compressed state distribution $p(\mathcal{Z}_{\text{train}})$ in tandem with the state distribution $p(\mathcal{X}_{\text{train}})$. Robust action inference is reached if during test time unseen states $\matr{x}$ can be reliably compressed into the same latent state space $p(\mathcal{Z}_{\text{train}})$ which was learned during the training procedure. Then, for this latent representation, a reliable translation to parameters for stabilizing linear Gaussian controllers can be expected. 



\noindent This leads to a policy behaviour, where inside $p(\mathcal{X}_{\text{train}})$, a GMR keeps the nonlinear generalization ability of neural networks and outside of $p(\mathcal{X}_{\text{train}})$, the linear stabilizing properties of linear Gaussian controllers are still being utilized. 

\subsection{Generative Motor Reflex Policy}
In order to enable these stabilizing properties, the compressed state space $\mathcal{Z}$ must keep the trained state space $\mathcal{X}_{\text{train}}$ sufficiently diverse and at the same time as compressed as possible. For this purpose, GMR exploits the idea of a variational autoencoder (VAE) \cite{vae}. A VAE consists of two concatenated networks: an encoder network, which maps an input to a latent representation and a decoder network, which retrieves an approximation of the original input from the latent representation (cf. Figure \ref{fig:gcm_model}). 

\noindent In GMR, the VAE enables to compress the robotic state $\matr{x}$ to a latent representation $\matr{z}$ while $\matr{z}$ is sufficiently diverse to reconstruct the robotic state $\matr{x}$ with the decoder network. By introducing noise in $\matr{z}$ during the S-step, the latent representation does not only represent the exact state $\matr{x}$ but also similar states. This increases the robustness of the reconstruction of the original state and as a consequence the robustness of the generated motor reflex parameters. \\
For retrieving $\matr{z}$, a mean $\bm{\mu}_z$ and a variance $\bm{\sigma}_z$ is determined first using an encoding neural network $h_{\text{enc}}$ under the weights $\bm{\Theta}_z$. 
\begin{align*}
[\bm{\mu}_z; \bm{\sigma}_z] = h_{\text{enc}}(\matr{x};\bm{\Theta}_z)
\end{align*}
Then, during training, the latent state $\matr{z}$ is retrieved by Gaussian sampling of $\bm{\mu}_z$ and $\bm{\sigma}_z$. This sampling step enables robustness in the estimation of $\bm{z}$ and leads to a robust translation of $\bm{z}$ to the motor reflex parameters $\bm{\Psi}$. Note that during test time, $\matr{z}$ is directly set to the mean value $\bm{\mu}_z$.
\begin{align*}
\matr{z} \sim \mathcal{N}(\bm{\mu}_z,\bm{\sigma}_z \odot \matr{I})
\end{align*}
Now, the state translation neural network $h_{\text{trans}}$ with weights $\bm{\Theta}_\Psi$ translates the latent representation $\matr{z}$ in motor reflex parameters $\bm{\Psi} = [\Psi_{\matr{K}}; \Psi_{\matr{k}}; \Psi_{\matr{\Sigma}}]$.
\begin{align*}
[\Psi_{\matr{K}}; \Psi_{\matr{k}}; \Psi_{\matr{\Sigma}}] = h_{\text{trans}}(\matr{z};\Theta_\Psi)
\end{align*}
The parameters $\bm{\Psi}$ are then used as parameters for a linear Gaussian controller (cf. Equation \ref{eq:lin_gaus_ctr}). By sampling from this controller, the action  $\matr{u}$ to be executed is finally retrieved.
\subsection{Generative Motor Reflex Training}
In Guided Policy Search, the policy weights $\bm{\Theta}_z$ and $\bm{\Theta}_\Psi$ are trained by supervised learning. For that, the C-step in GPS provides a dataset $\mathcal{D}$ which are tuples of the form $(\matr{x}; \bm{\Psi}) \subseteq \mathcal{D}$. Using this dataset, a variant of gradient descent is used to optimize the weights by minimizing the loss term, denoted as $\mathcal{L}_{\text{GMR}}$ (cf. Figure \ref{fig:gcm_model}).\\
The loss term has to enforce a compressed and meaningful latent state representation while the motor reflex parameters are generated with high precision. Therefore, we propose a loss term which consists of four parts:
(i) minimizing the mean-squared state reconstruction loss, (ii) the KL divergence from a unit distribution to the latent state distribution, (iii) the KL divergence prediction error of the motor reflex parameters and (iv) an L2 regularization term. Therein, the sum of (i) and (ii) is the standard loss function of a variational autoencoder, where (ii) forces the compression of the state space to a latent state space which is in VAE a unit Gaussian distribution. The prediction error (iii) 
trains the translation model for generating motor reflex parameters and (iv) is a regularization term that improves overall stability and is already used by previous GPS policies.
\begin{align}
\label{eq:gmr_loss}
\mathcal{L}_{\text{GMR}} = \frac{1}{|\mathcal{D}|} &\sum_{\matr{x}, \bm{\Psi} \in \mathcal{D}} \Big( \underbrace{||\matr{x}-h_{\text{dec}}(\matr{z};\bm{\Theta}_z)||_2^2}_{\text{(i)}}  \nonumber \\ &+ \alpha  \underbrace{D_{\text{KL}}(\matr{z}|| \mathcal{N}(0,I))}_{\text{(ii)}} \nonumber \\ &+ \underbrace{ D_{\text{KL}}(\pi_{\Psi}(\matr{u}|\matr{x})||p_{i}(\matr{u}|\matr{x}))}_{\text{(iii)}} \Big)  + \beta \underbrace{\sum \bm{\Theta}^2}_{\text{(iv)}} 
\end{align} 
In this loss term $\beta$ weights the influence of the L2 regularization and $\alpha$ the influence of the KL divergence on the overall loss. A greater value for $\alpha$ results in a more compressed latent space, but reduces the ability to represent the state space exactly. At the current state of work, $\alpha$ is hand tuned.

\subsection{Generative Motor Reflex with Guided Policy Search}
\begin{algorithm}[b]
\caption{Mirror Descent GPS with GMR}\label{alg:mdgpsgmr_algorithm}
\begin{algorithmic}[1]
\For{$n$-iteration $n=1$ to $N$}
\State Generate samples $\mathcal{T}=\{\tau_i\}$ by running $p_i$ or $\pi$
\State Fit Gaussian dynamics $p_i(\matr{x}_{t+1}|\matr{x}_t, \matr{u}_t)$
\Do
\State \textbf{C1-step}: LQR backward $\mathcal{D} \leftarrow [\Psi_{\matr{K}i}; \Psi_{\matr{k}i}; \Psi_{\matr{\Sigma}i}]$
\State \textbf{C2-step}: LQR forward $\mathcal{D} \leftarrow \matr{x}_i$
\State \textbf{C3-step}: Adjust dual variable $\lambda$
\doWhile{$D_{\text{KL}}(p_{i}(\tau)||\pi(\tau)) - \epsilon > 0$}
\State \textbf{S-step}: $\pi \leftarrow \argmin_{\bm{\Theta}_z,\bm{\Theta}_x,\bm{\Theta}_{\Psi}} \mathcal{L}_{\text{GMR}}(\mathcal{D}, \alpha)$ 
\EndFor
\end{algorithmic}
\end{algorithm}
Algorithm \ref{alg:mdgpsgmr_algorithm} presents the adapted Policy Search method for training GMR, where we chose a GPS variant, which is the \textit{Mirror Descent GPS} (MDGPS) \cite{c33}. Other variants of GPS can be adapted analoguesly. \\ 
In MDGPS, the C-step minimizes the following Lagrangian loss term:
\begin{equation*}
\mathcal{L}_{\text{GPS}}= \underbrace{J(\tau)}_{\text{(i)}}+  \underbrace{\sum_{t=1}^T \lambda _t D_{\text{KL}}(p_{i}(\tau)||\pi(\tau))}_{\text{(ii)}}
\end{equation*}
This loss term consists of two parts: (i) the trajectory loss term $J(\tau)$ and (ii) the constraint $D_{\text{KL}}(p_{i}(\tau)||\pi(\tau)) \leq \epsilon$, which bounds the change in the optimized trajectory distribution $p_i(\tau)$ and by utilizing 
by the Lagrange multiplier $\lambda$. Now, the loss term $\mathcal{L}_{\text{GPS}}$ is minimized using dual gradient descent (DGD). \\
DGD is an optimization approach, which enables to incorporate Lagrange multipliers. For DGD in GPS, three steps are performed: Given an initial $\lambda$, trajectory optimization is utilized for minimizing $\mathcal{L}_{\text{GPS}}$ (C1- and C2-step). Then, the dual variable $\lambda$ is updated stepwise until the constraint violation 
$D_{\text{KL}}(p_{i}(\tau)||\pi(\tau)) - \epsilon$ (C3-step) is met. In prior GPS methods, the output of this minimization process is an optimized trajectory distribution $p_i(\tau)$. 
However, for training GMR we need a dataset $\mathcal{D}$ with pairs of states $\matr{x}$ and corresponding motor reflex parameters $\bm{\Psi}$. 
This dataset can be retrieved during the C1 and C2-step.

\noindent Therein, the optimized trajectory distribution is computed by a \textit{linear-quadratic regulator} (LQR). An LQR consists of a backward 
(C1-step) and a forward pass (C2-step).
Within the LQR backward pass, a value function is minimized that represents the accumulated cost starting from the state $\matr{x}_t$ to goal state $\matr{x}_T$. 
\begin{equation*} \label{eq:cost_to_go}
V(\matr{x}_t)=\min_{\matr{u}_t}Q_t(\matr{x}_t,\matr{u}_t)=\min_{\matr{u}_t} \mathcal{L}_{\text{GPS}}(\matr{x}_t,\matr{u}_t)+E[V(\matr{x}_{t+1})]
\end{equation*}
With having computed an approximation of the Q-function by a second-order Taylor expansion, the optimal action can be found by the partial derivation with respect to $\matr{u}_t$:
\begin{equation}
\label{eq:lqr_ctr}
\matr{u}_t=\argmin_{\matr{u}_t}Q_t(\matr{x}_t,\matr{u}_t)=-Q_{\matr{u},\matr{u}t}^{-1}(Q_{\matr{u}t}+Q_{\matr{u},\matr{x}t}\matr{x}_t)
\end{equation}
Now, Equation \ref{eq:lqr_ctr} can be transformed to a linear Gaussian controller by substituting the following motor reflex para-meters in timestep $t$:
\begin{align*}
\Psi_{\matr{K}t}&=-Q_{\matr{u},\matr{u}t}^{-1}Q_{\matr{u},\matr{x}t} \\
\Psi_{\matr{k}t}&=-Q_{\matr{u},\matr{u}t}^{-1}Q_{\matr{u}t} \\
\Psi_{\bm{\Sigma}t}&=Q_{\matr{u},\matr{u}t}^{-1}
\end{align*}
As proposed in previous GPS approaches \cite{c33}, we set the motor reflex covariance $\Psi_{\bm{\Sigma}t}$ to the inverse of the action related Q-value $Q_{\matr{u},\matr{u}t}^{-1}$. The intuition  
behind this is to keep the exploration noise $\Psi_{\bm{\Sigma}t}$ low in case that the actions change the Q-value significantly. 
Now, the first and second derivatives of the value function can be derived as follows:
\begin{align*}
Q_{\matr{xu},\matr{xu}t}&=l_{\matr{xu},\matr{xu}t}+f_{\matr{xu}t}^T V_{\matr{x},\matr{x}t+1} f_{\matr{xu}t} \\
Q_{\matr{xu}t}&=l_{\matr{xu}t}+f_{\matr{xu}t}^T V_{\matr{x}t+1} \\
V_{\matr{x},\matr{x}t}&=Q_{\matr{x},\matr{x}t}+Q_{\matr{u},\matr{x}t}\Psi_{\matr{K}t} \\
V_{\matr{x},t}&=Q_{\matr{x}t}+Q_{\matr{u},\matr{x}t}\Psi_{\matr{k}t}
\end{align*}
Using these equations, the LQR backward pass calculates $\Psi_{\matr{K}t}, \Psi_{\matr{k}t}$ and $\Psi_{\bm{\Sigma}t}$ as well as the Q- and V-values for each timestep. 
Then, in a following forward pass (C2-step) we exploit the learned dynamics to retrieve the corresponding states $\matr{x}_t$ for $\Psi_{\matr{K}t}, \Psi_{\matr{k}t}, \Psi_{\bm{\Sigma}t}$ given an initial state $\matr{x}_0$.
After the C-steps, the GMR weights $\bm{\Theta}_z$, $\bm{\Theta}_x$, $\bm{\Theta}_{\Psi}$ are trained within the S-step. This minimizes the loss function presented in Equation \ref{eq:gmr_loss} and leads to the final GMR policy.
\section{Experiments}
\begin{figure}[bt]
	\centering
	\includegraphics[width=0.4\textwidth]{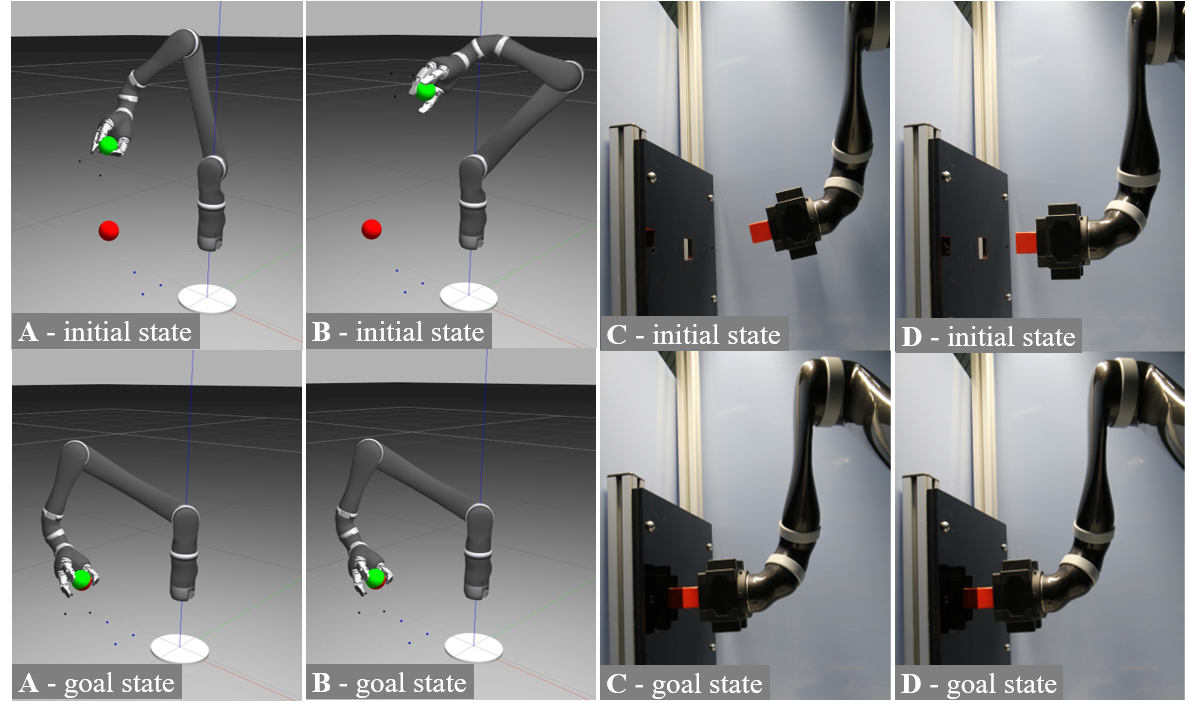}
    \caption{Evaluation scenarios for the presented Generative Motor Reflex Policy. Scenario A and B are two conditions of a reaching task that share the same target state and different initial states; scenario C and D are two conditions of a (wooden) block in hole scenario, which has a light press fit.}
    	\label{fig:all_scenarios}
\end{figure}

The proposed learning algorithm is evaluated by using a six axes robot arm of the type \textit{Kinova Jaco 2} for simulated and real world manipulation tasks (cf. Figure \ref{fig:all_scenarios}).
Within the scope of this work the object 
to be manipulated is rigidly connected to the gripper. The implementation is built upon the guided policy search toolbox \cite{c32} and is utilizing Gazebo for simulation.  

\noindent We evaluate two properties of our approach: (i) the learning performance and (ii) the robustness of GMR. 
The evaluation scenarios include two conditions of simulated reaching tasks (cf. Figure \ref{fig:all_scenarios}-A, \ref{fig:all_scenarios}-B) and two conditions of real-world peg-in-hole tasks with a light press fit (cf. Figure \ref{fig:all_scenarios}-C, \ref{fig:all_scenarios}-D).

\noindent In each iteration, the robot collects five samples per condition with a length of $T=80$ timesteps. 
We trained the robot in each experiment for $N=10$ iterations. This process leads to a total of $50$ trajectory samples for each condition. 
The continuous state space of the robot consists of six joint positions and six joint velocities. The action space includes six joint torques that are controlled by the policy with $20\,$Hz.
This leads to a parameter space of the motor reflex with $114$ dimensions ($\Psi_{\matr{K}} \in \mathbb{R}^{6\times12}, \Psi_{\matr{k}} \in \mathbb{R}^{6}, \Psi_{\matr{\Sigma}} \in \mathbb{R}^{6\times6}$). \\ 
We set the L2 regularization term to $\beta=0.0002$. 
The weights $\bm{\Theta}$ of the policy topology presented in Figure \ref{fig:gcm_model} are optimized by the \textit{ADAM} 
solver, whereby the number of training epochs was set to $400$ and the batch size to $20$. 
We compared our policy with the prior state-action policy trained by MDGPS. Here we used the same settings for the policy training, except for the difference in the topology that was set to a fully connected neural network with two hidden layers with $64$ rectified linear units. The supplementary video compares the resulting motion behaviour of GMR and MDGPS policies\footnote{\texttt{https://philippente.github.io/pub/GMR.html}}.

\subsection{Learning Performance}
\begin{figure}[bt]
\subfigure[Trained on cond. A]{\includegraphics[height=3.1cm, width=0.24\textwidth, trim=0 3 0 0,clip]{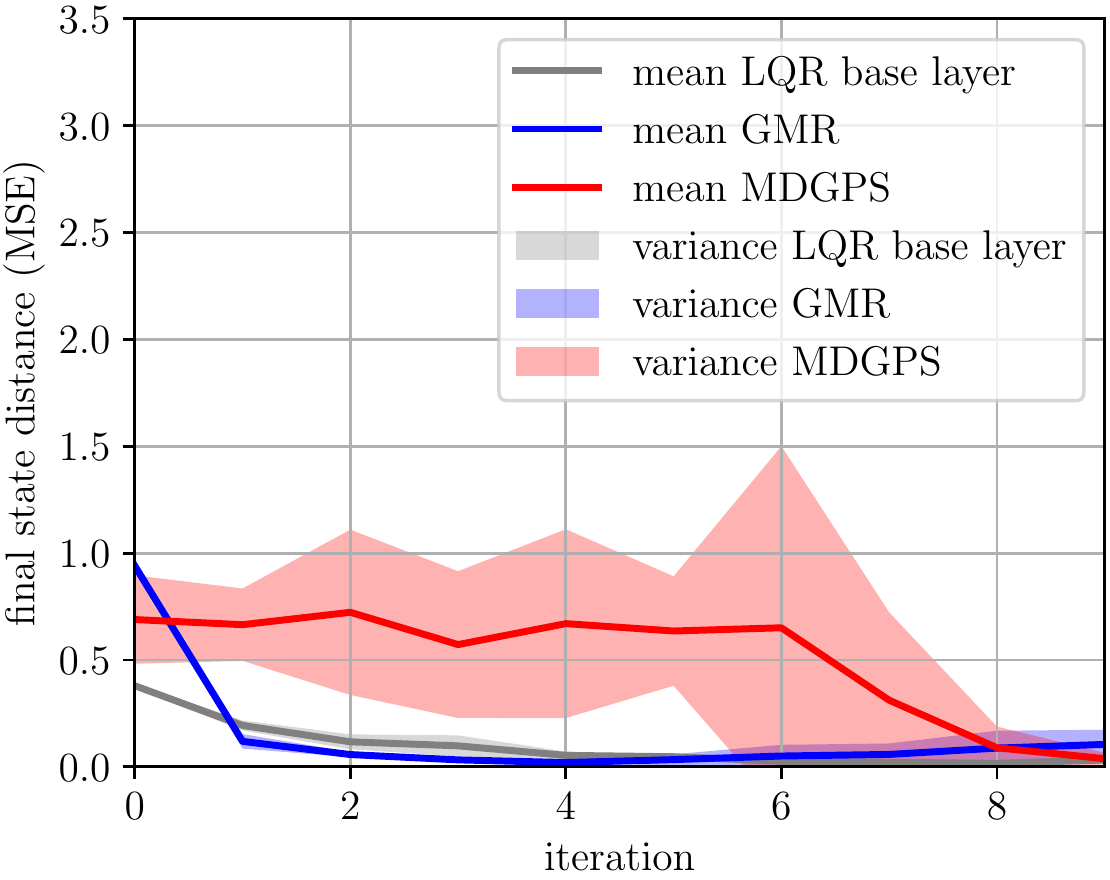}}
\subfigure[Trained on cond. A and B]{\includegraphics[height=3.1cm,width=0.24\textwidth, trim=0 3 0 0,clip]{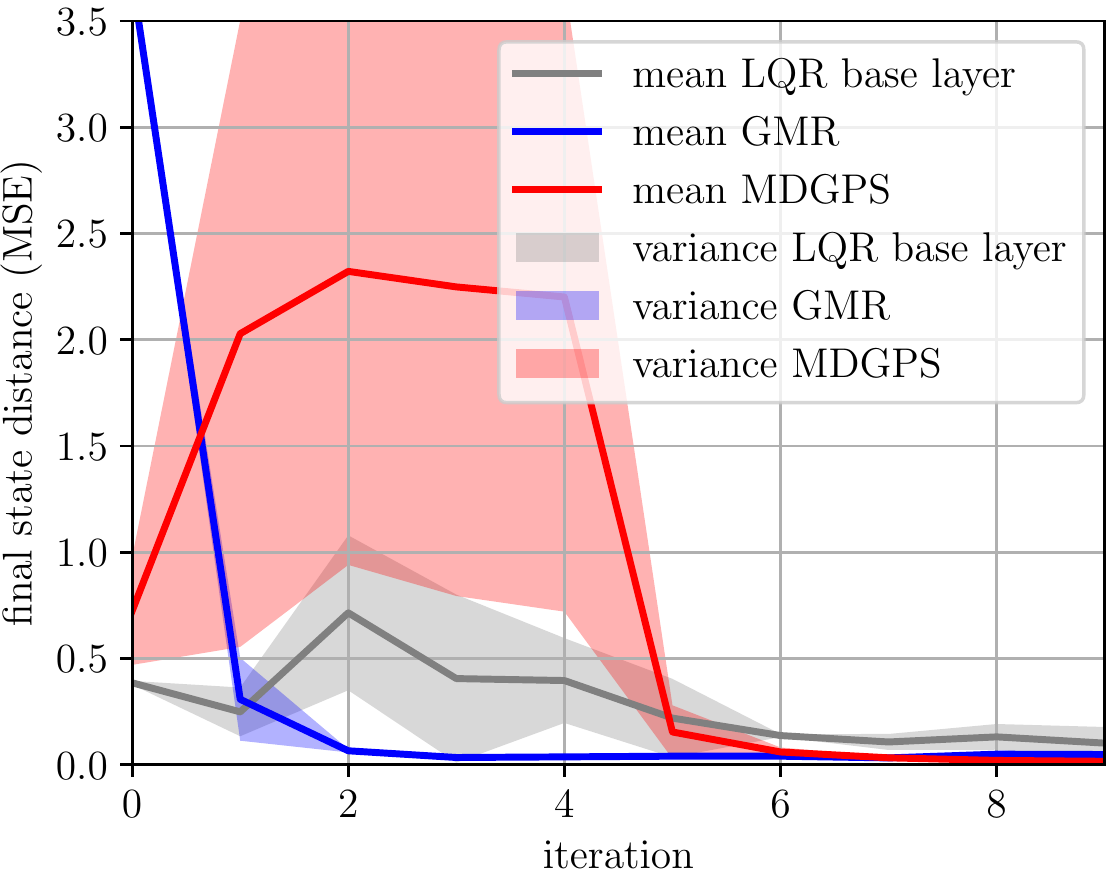}}
\caption{Average performance of the MDGPS and GMR policy measured by the mean-squared-error (MSE) of the final state distance. All experiments are repeated six times for each algorithm, with the MSE and variance shown in the plots. In addition to MDGPS and GMR, we add the average performance of the linear Gaussian controller (denoted as LQR), which serves as a base layer for training MDGPS and GMR policies in each iteration.}
\label{fig:mse_accuracy}
\end{figure}

First, we evaluated the learning performance of the GMR training compared to MDGPS and the linear Gaussian controller base layer, where only the C-step is performed (hereinafter LQR base layer). In Figure \ref{fig:mse_accuracy} the mean squared error (MSE) of the final state distance for the reaching task in case of one and two conditions is compared between GMR, MDGPS and the LQR base layer. The MSE describes the distance in the final joint states velocities. 

\noindent During training time of GMR policies, new LQR base layers are computed after each iteration, which are then used to continue the supervised training of the GMR policies. For both training scenarios (cf. Figure \ref{fig:all_scenarios}), GMR requires two iterations to imitate the LQR base layer reliably. In contrast to the expectations, the GMR solution outperforms the LQR base layer after a few iterations. This property is a result of the generalization ability of the GMR, which is learned by training the policy on two LQR base layers of successive iterations. 

\noindent The learning progress of GMR shows a small variance in the performance, which points out the reliability of our approach. 
Compared to this, the performance of the MDGPS policy has a high variance during training and usually needs $2$ - $4$ times more iterations for a successful imitation of the LQR base layer. 

\subsection{Robustness}
\begin{figure}
\subfigure[GMR trained with cond. A]{\includegraphics[height=3.35cm,width=0.24\textwidth, trim=65 4 1 6,clip]{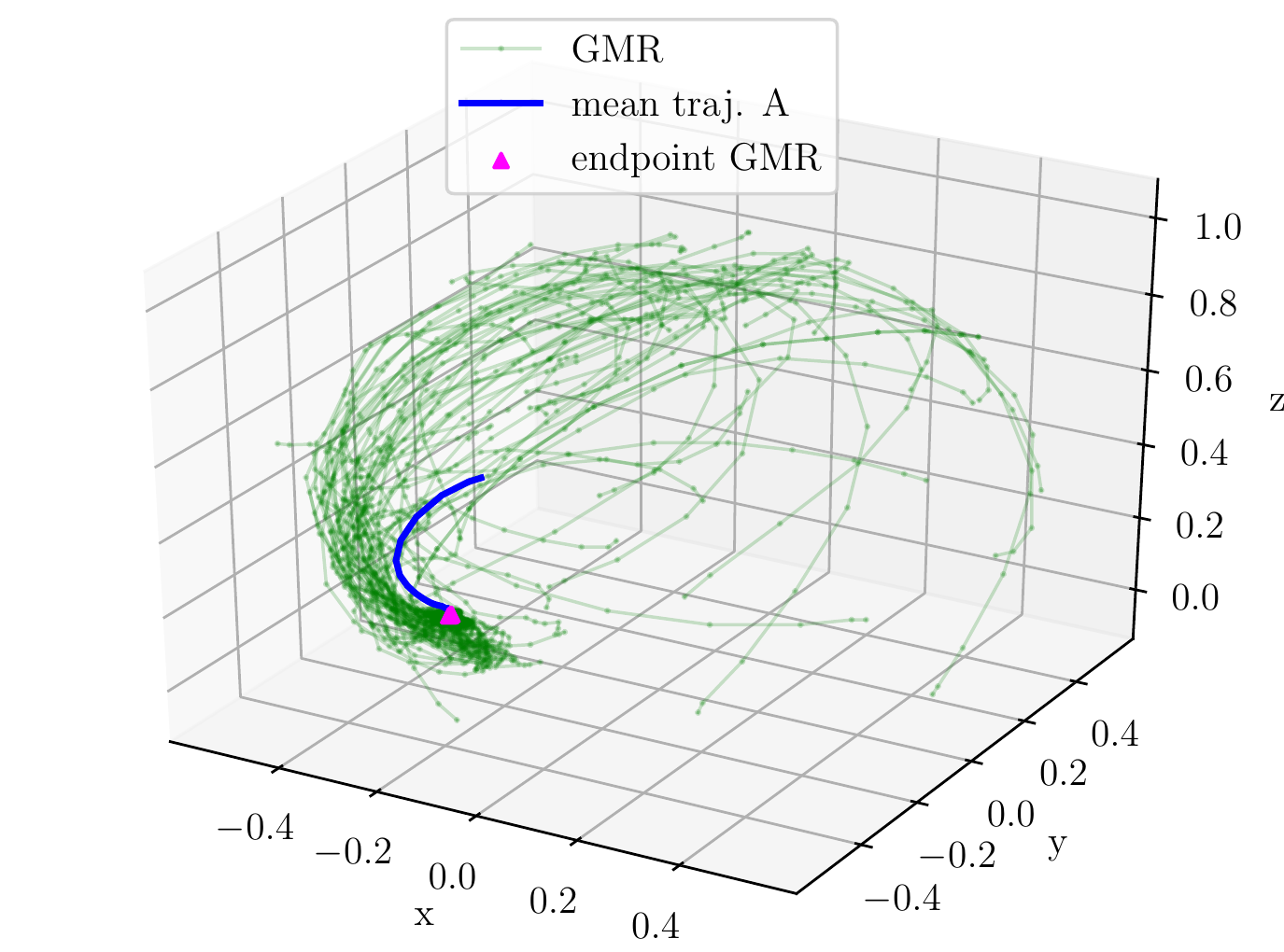}}
\subfigure[GMR trained with cond. A, B]{\includegraphics[height=3.35cm,width=0.24\textwidth, trim=65 4 1 6,clip]{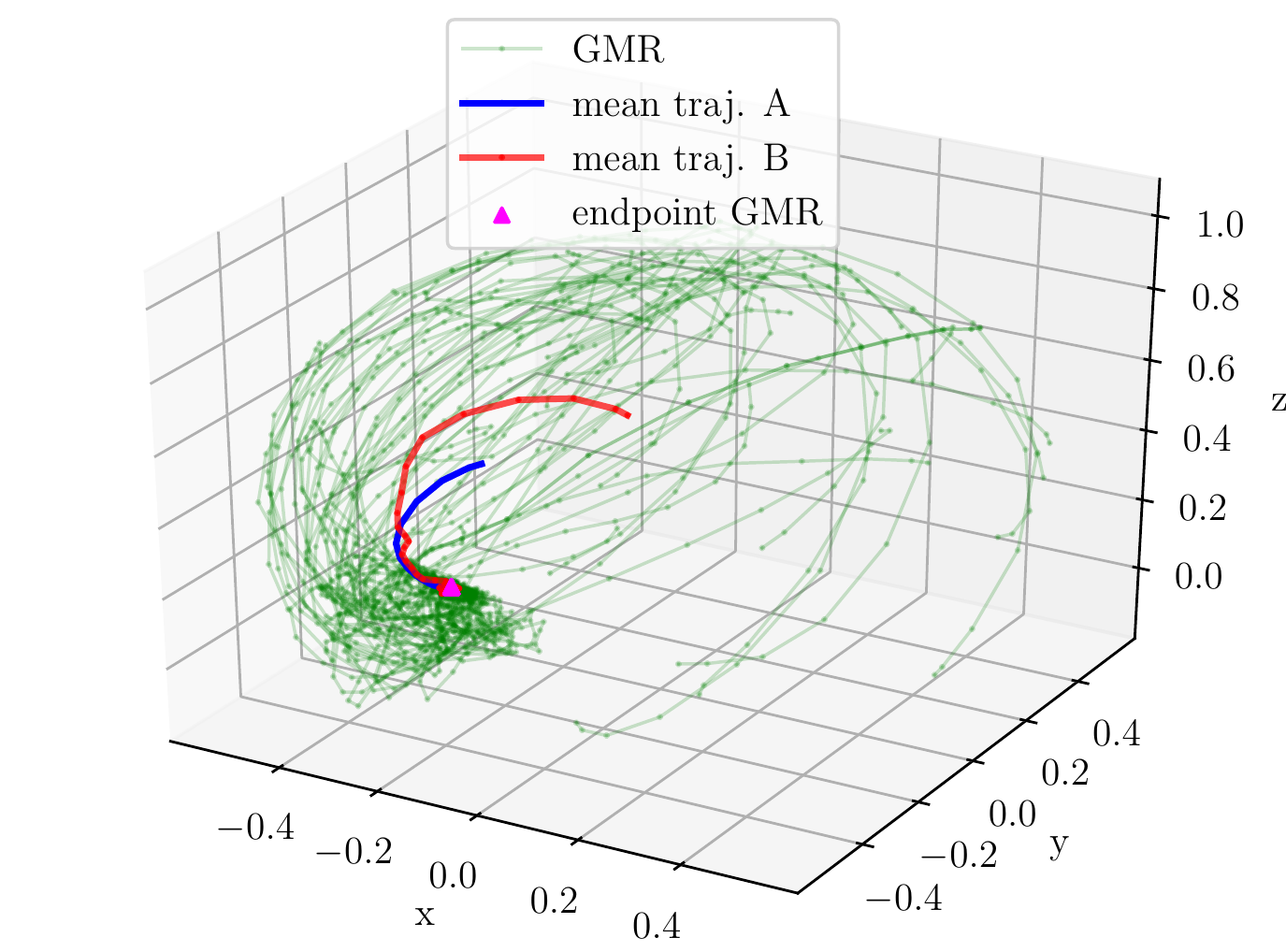}}
\subfigure[MDGPS trained with cond. A]{\adjincludegraphics[height=3.35cm,width=0.24\textwidth, trim=65 4 1 6,clip]{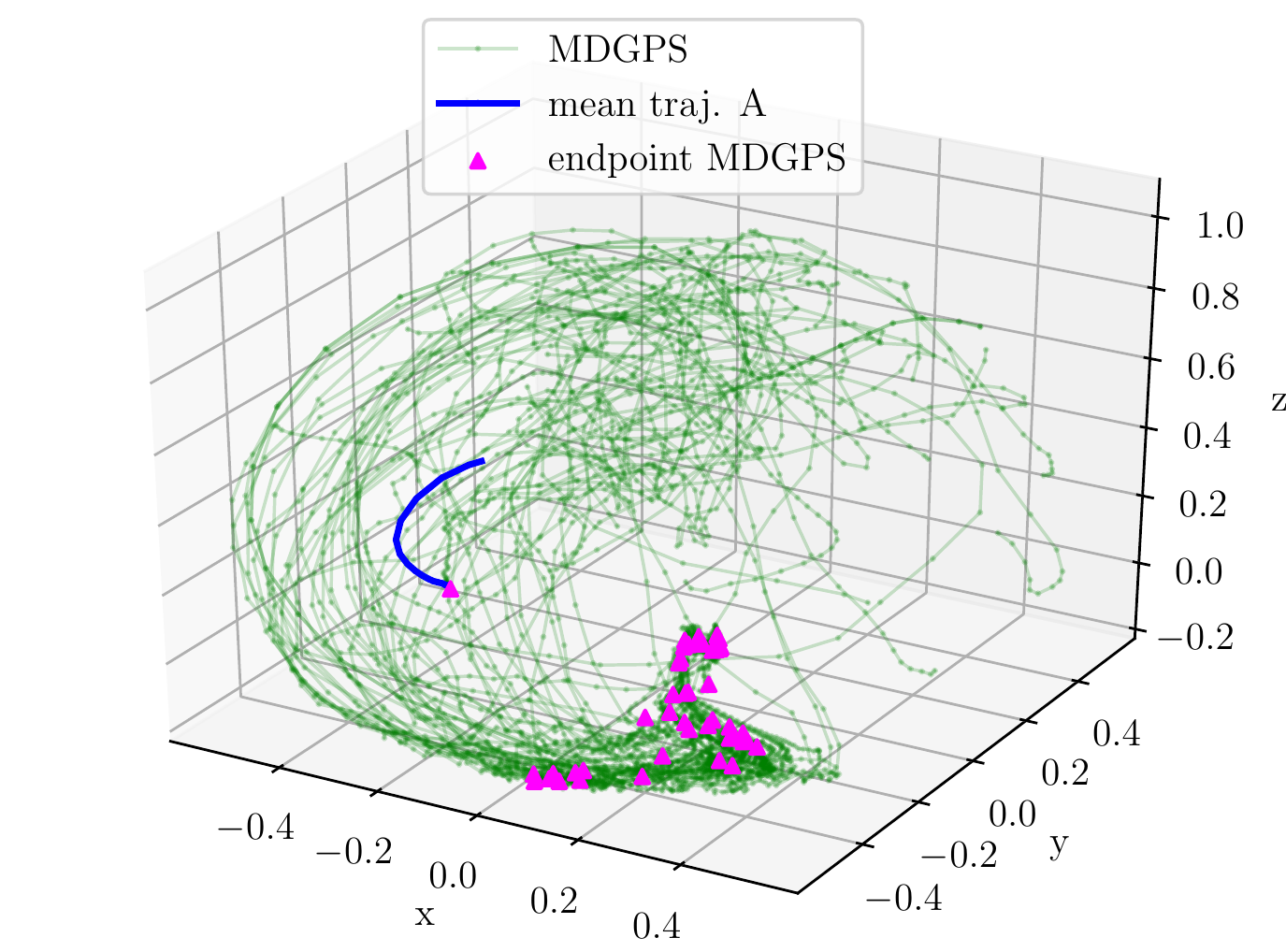}}
\subfigure[MDGPS trained with cond. A, B]{\adjincludegraphics[height=3.35cm,width=0.24\textwidth, trim=65 4 1 6,clip]{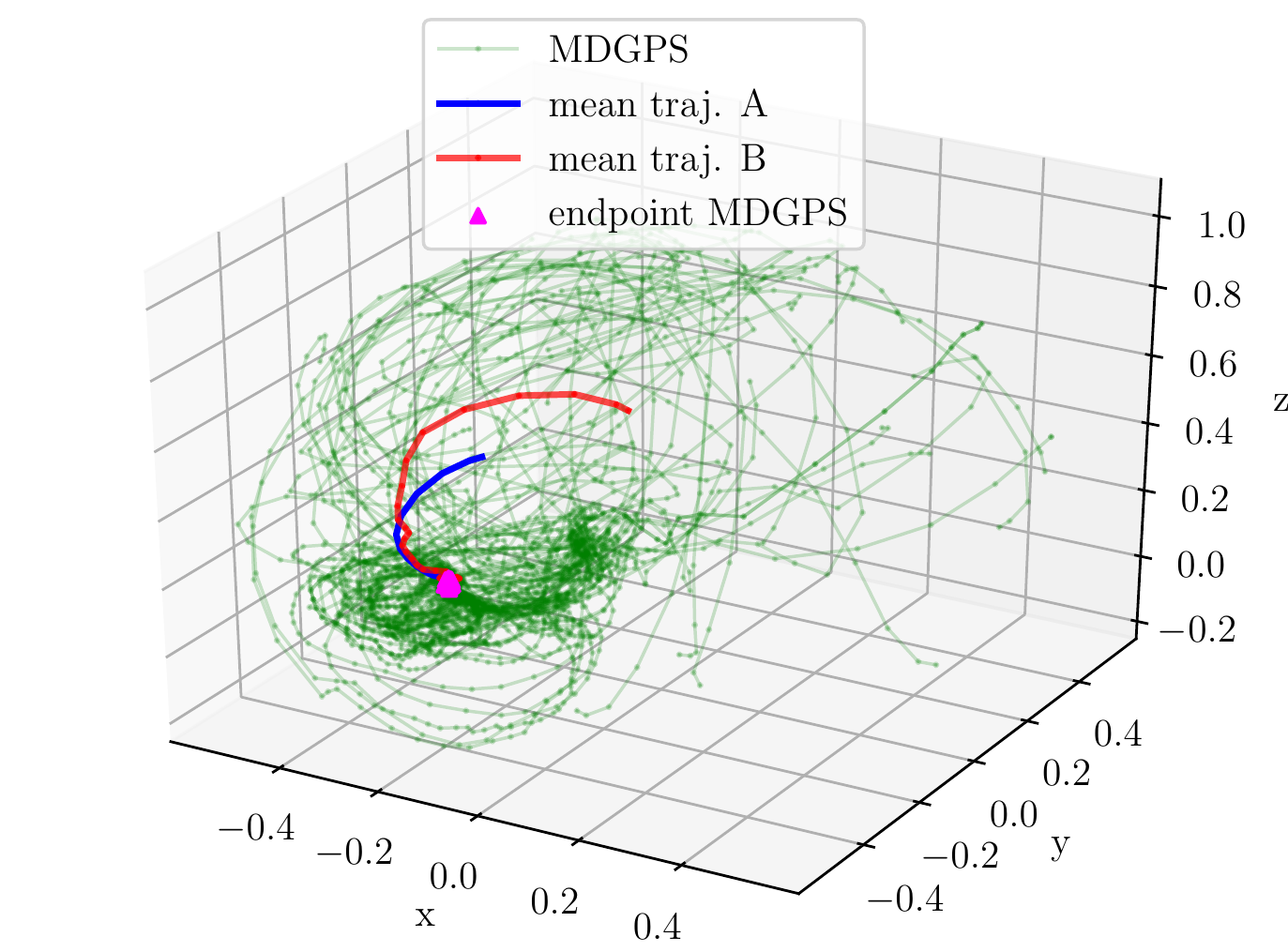}}
\caption{Trajectories produced by MDGPS and GMR policies for 50 random initial states. Here, the policy was trained for N=10 iterations, whereas the robot collected in each iteration five samples. For each scenario, the mean trajectory executed by the robot during the exploration phase is plotted. The endpoints show the final state reached by the policies.}
\label{fig:robustness}
\end{figure}

Robustness to unseen states is the key advantage of GMR. We show, that GMR produces stable motions over a much broader state space compared to MDGPS. First, we trained the simulated robot on scenario A and B (cf. Figure \ref{fig:all_scenarios}). Then, we set the initial robot state to $50$ random, uniform distributed configurations over the whole state space. 
From there, the robot executed the policy, which is either a GMR or MDGPS policy.
The diagram in Figure \ref{fig:robustness} shows the resulting trajectories of the end effector for both policies. Table \ref{table:suc_rate_gen} provides an overview of the success rate for reaching the finale state. In our experiments we observed that a $\text{MSE} < 0.1$ can be interpreted as successful. 
\begin{table}[bt]
\caption{The success rate of MDGPS and GMR policies for reaching the final state (MSE $< 0.01$) after $N=10$ iterations for 50 random initial states. The supplementary video gives an impression of both policy performances.}
\label{table:suc_rate_gen}
\begin{center}
\begin{tabular}{|c||c||c||c||c|}
\hline
training scenario & MDGPS & GMR\\
\hline
A & 12/50 & 50/50\\
A,B & 50/50 & 50/50\\
C & 4/50 & 39/50\\
C,D & 9/50 & 44/50\\
\hline
\end{tabular}
\end{center}
\end{table}
\noindent For the simulated scenarios A and B, GMR reaches the final state for all random initial states.  
In comparison, the MDGPS policy fails in $38$ of $50$ trials in case of one training condition. Within the simulated scenario, this problem can be solved by using more training conditions (which requires more training time). Nevertheless, we usually observe an extreme overshooting at the end of the motions for all scenarios, cf. Figure \ref{fig:robustness}-c and \ref{fig:robustness}-d. 

\noindent However, in the real-world scenarios C and D the high failure rate remains even for two training conditions. Usually, the MDGPS policy starts to oscillate even by small state disturbances. If the state disturbance increases, the MDGPS policy diverged from the goal state, whereas the GMR policy typically reaches the goal state with less overshooting. 

\noindent In conclusion, we can state that GMR policies are an approach to reach robustness outside the explored trajectory distribution even with a low number of training conditions.

\section{Conclusion and future work}
We could improve the robustness of neural network policies by using motor reflexes as an intermediate step and evaluated our approach for manipulation tasks. Motor reflexes are one time step controller with local stabilizing properties. 



\noindent So far, these motor reflexes are linear Gaussian controllers. In future work, we will research also other motor reflex representations, i.e. error minimizing dynamic systems like Dynamic Movement Primitives. In addition, a promising direction for further research is the embedding of vision-based feature extractors for a more general usage of GMR. 

\noindent However, in terms of the robustness of GMR, stability analysis could be researched. A key property of generative models is the ability to generate samples from a latent space, which are in GMR pairs of states and motor reflex parameters. This is not tackled in this work yet, but could potentially be an approach for validating GMR.

\section*{ACKNOWLEDGMENT}

This contribution is funded by VDMA in germany as part of the project "InPulS - Intelligent processes and learning systems" of the Forschungskuratorium Maschinenbau (FKM).   

\addtolength{\textheight}{0cm}   



\section{Appendix}
\subsection{KL-divergence in GMR loss function}
The KL term (iii) in Equation \ref{eq:gmr_loss} minimizes the divergence between two conditional Gaussian's. For that reason, this term can be rewritten by exploiting the definition of the KL-divergence for two Gaussian's as:
\begin{align*}
D_{\text{KL}}(\pi_{\Psi}(\matr{u}|\matr{x}) || p_i(\matr{u}|\matr{x}))=\mathcal{L}_{\text{cov}} + \mathcal{L}_{\text{mean}}
\end{align*}
where 
\begin{align*}
\mathcal{L}_{\text{cov}} = \text{tr}[\matr{\Sigma}_{p_i}^{-1} \matr{\Psi_{\Sigma}}(\matr{x})] - \log |\matr{\Psi_{\Sigma}}(\matr{x})|
\end{align*}
and 
\begin{align*}
    \mathcal{L}_{\text{mean}} &= 
    (\Delta\matr{u})\matr{\Sigma}_{p_i}^{-1}(\Delta\matr{u}) \\
    &= (\Delta\matr{K}\matr{x} - \Delta \matr{k})\matr{\Sigma}_{p_i}^{-1}(\Delta\matr{K}\matr{x} - \Delta \matr{k})
\end{align*}
with
\begin{align}
    \Delta\matr{u} &= \matr{u}_{\Psi} - \matr{u}_{p_i} \\
    \Delta\matr{K} &= \matr{\Psi}_{\matr{K}} - \matr{K}_{p_i} \\
    \Delta\matr{k} &= \matr{\Psi}_{\matr{k}} - \matr{k}_{p_i} \label{eq:appendix_k}
\end{align}
This loss term minimizes the KL-divergence of the controller distribution $p(\matr{u}|\matr{x})$ to the controller distribution generated by GMR, which is $\pi(\matr{u}|\matr{x})$. However, we observed that the controller parameters $\matr{K}_{p_i}$ and $\matr{k}_{p_i}$ of $p(\tau)$ tend to overfit in cases, where the same state is visited multiple times. This situation occurs especially often in the neighbourhood of the initial state and the goal state.  

\noindent In prior GPS methods, this overfitting is handled by training the policy $\pi$ on the inferred actions $\matr{u}$ instead of controller parameters. However, GMR aims to train on a linear-feedback policy which is described by $\matr{\Psi_K}, \matr{\Psi_k}$. In doing so, little training errors of $\matr{\Psi_K}, \matr{\Psi_k}$ can lead to strongly deviating actions $\matr{u}$.

\noindent For that reason, we propose a training on $\matr{\Psi_K}$ and corresponding actions $\matr{u_{p_i}}$ coming from the original LQR policy $p_i$. From this training values, the offset $\matr{\Psi_k}$ follows indirectly, which improves the overall accuracy of the final action $\matr{u}$. This more robust training procedure is reached by replacing $\Delta\matr{k}$ in Equation \ref{eq:appendix_k} with the following expression.
\begin{align*}
    \Delta\matr{k} = \Delta\matr{u} - \Delta\matr{K}\matr{x} .
\end{align*}



\begin{thebibliography}{99}
\bibitem{c2} Levine, S. and Abbeel P. "Learning neural network policies with guided policy search under unknown dynamics." Advances in Neural Information Processing Systems. 2014.
\bibitem{c6} Endo, G., Morimoto, J., Matsubara, T., Nakanishi, J. and Cheng, G. "Learning CPG-based biped locomotion with a policy gradient method: Application to a humanoid robot." The International Journal of Robotics Research 27.2 (2008): 213-228.
\bibitem{c7} Peters, J. and Schaal, S. "Reinforcement learning of motor skills with policy gradients." Neural networks 21.4 (2008): 682-697.
\bibitem{c8} Levine, S., Wagener, N., and Abbeel, P. "Learning contact-rich manipulation skills with guided policy search." Robotics and Automation (ICRA), 2015 IEEE International Conference on (pp. 156-163).
\bibitem{c9} Theodorou, E., Buchli, J. and Schaal, S. "Reinforcement learning of motor skills in high dimensions: A path integral approach." Robotics and Automation (ICRA), 2010 IEEE International Conference on. IEEE, 2010.
\bibitem{c5} Kohl, N. and Stone, P. "Policy gradient reinforcement learning for fast quadrupedal locomotion." Robotics and Automation, 2004. Proceedings. ICRA'04. 2004 IEEE International Conference on. Vol. 3. IEEE, 2004.
\bibitem{c1} Deisenroth, M. P., Neumann, G. and Peters, J. "A survey on policy search for robotics." Foundations and Trends® in Robotics 2.1–2 (2013): 1-142.
\bibitem{c10} Peters, J., M{\"u}lling, K. and Altun, Y. "Relative Entropy Policy Search." AAAI. 2010.
\bibitem{c11} Kalakrishnan, M., Righetti, L., Pastor, P. and Schaal, S. "Learning force control policies for compliant manipulation." Intelligent Robots and Systems (IROS), 2011 IEEE/RSJ International Conference on. IEEE, 2011.
\bibitem{c12} Inoue, T., De Magistris, G., Munawar, A., Yokoya, T. and Tachibana, R. Deep reinforcement learning for high precision assembly tasks. In Intelligent Robots and Systems (IROS), 2017 IEEE/RSJ International Conference on (pp. 819-825).
\bibitem{c13} Abbeel, P., Coates, A., Quigley, M. and Ng, A. Y. An application of reinforcement learning to aerobatic helicopter flight. In Advances in neural information processing systems (2007): 1-8.
\bibitem{c14} Gu, S., Holly, E., Lillicrap, T. and Levine, S. "Deep reinforcement learning for robotic manipulation with asynchronous off-policy updates." Robotics and Automation (ICRA), 2017 IEEE International Conference on. IEEE, 2017.
\bibitem{c14.1} Schaal, S. "Dynamic movement primitives-a framework for motor control in humans and humanoid robotics." Adaptive motion of animals and machines. Springer, Tokyo, 2006. 261-280.
\bibitem{c14.1.2} A. Rai, G. Sutanto, S. Schaal, and F. Meier, “Learning feedback terms for reactive planning and control,” in Robotics and Automation (ICRA), 2017 IEEE International Conference on, 2017, pp. 2184–2191.
\bibitem{c14.1.3} G. Sutanto, Z. Su, S. Schaal, and F. Meier, “Learning sensor feedback models from demonstrations via phase-modulated neural networks,” in 2018 IEEE International Conference on Robotics and Automation (ICRA), 2018, pp. 1142–1149.
\bibitem{c14.2} Paraschos, A., Daniel, C., Peters, J. R. and Neumann, G. "Probabilistic movement primitives." In Advances in neural information processing systems (2013): 2616-2624.
\bibitem{c14.2GMM} Calinon, S., Guenter, F. and Billard, A. On learning, representing, and generalizing a task in a humanoid robot. IEEE Transactions on Systems, Man, and Cybernetics, Part B (Cybernetics), 37(2), (2007): 286-298.
\bibitem{14.2HMM} Billard, A., Calinon, S., Dillmann, R. and Schaal, S. Robot programming by demonstration. In Springer handbook of robotics (2008): 1371-1394. Springer, Berlin, Heidelberg.
\bibitem{14.2ANN} Levine, S., Finn, C., Darrell, T. and Abbeel, P. End-to-end training of deep visuomotor policies. The Journal of Machine Learning Research, 17(1), (2016): 1334-1373.
\bibitem{c14.25} Morimoto, J. and Kenji D. "Robust reinforcement learning." Neural computation 17.2 (2005): 335-359.
\bibitem{c14.3} Pinto, L., Davidson, J., Sukthankar, R. and Gupta, A. "Robust adversarial reinforcement learning." arXiv preprint arXiv:1703.02702 (2017).
\bibitem{c14.4} Pattanaik, A., Tang, Z., Liu, S., Bommannan, G., Chowdhary, G. "Robust deep reinforcement learning with adversarial attacks." In Proceedings of the 17th International Conference on Autonomous Agents and MultiAgent Systems (2018): 2040-2042. International Foundation for Autonomous Agents and Multiagent Systems.
\bibitem{vae} Kingma, D. P., and Welling, M. Auto-encoding variational bayes. arXiv preprint arXiv:1312.6114 (2013).
\bibitem{c33} Montgomery, W. H. and Levine, S. "Guided policy search via approximate mirror descent." Advances in Neural Information Processing Systems. 2016.
\bibitem{c32} Finn, C., Zhang, M., Fu, J., Montgomery, W., Tan, X.Y., McCarthy, Z., Stadie, B., Scharff, E. and Levine, S. Guided Policy Search Code Implementation. 2016. Software available from rll.berkeley.edu/gps.




\end{thebibliography}
\end{document}